\documentclass[lettersize,journal]{IEEEtran}
\usepackage{amsmath,amsfonts}
\usepackage{algorithmic}
\usepackage{array}
\usepackage{textcomp}
\usepackage{stfloats}
\usepackage{url}
\usepackage{verbatim}
\usepackage{graphicx}
\usepackage[pagebackref,breaklinks,colorlinks,citecolor=blue,linkcolor=red,urlcolor=blue]{hyperref}
\usepackage{booktabs}
\usepackage{multirow}   

\usepackage{siunitx}
\usepackage[most]{tcolorbox}
\newtcolorbox{findingsbox}[2][]{%
  colback=blue!3,
  colframe=blue!60!black,
  coltitle=black,
  fonttitle=\bfseries,
  colbacktitle=blue!15,
  boxrule=0.8pt,
  arc=2mm,
  top=1.5mm, bottom=1.5mm,
  left=1.5mm, right=1.5mm,           
  enhanced,
  attach boxed title to top left={xshift=2mm,yshift=-2mm},
  #1
}

\usepackage{array}
\newcolumntype{P}[1]{>{\centering\arraybackslash}m{#1}}

\usepackage[table]{xcolor}
\definecolor{iccvblue}{rgb}{0.21,0.49,0.74}
\usepackage{balance}
\usepackage{orcidlink}

\usepackage[caption=false,font=normalsize,labelfont=sf,textfont=sf]{subfig}

\begin{document}

\title{Towards Comprehensive Interactive Change Understanding in Remote Sensing: A Large-scale Dataset and Dual-granularity Enhanced VLM}

\author{Junxiao Xue~\orcidlink{0000-0003-1569-5362}, Quan Deng~\orcidlink{0009-0008-2401-2432}, Xuecheng Wu~\orcidlink{0000-0002-6244-0269}, Kelu Yao~\orcidlink{0000-0002-4891-3197}, Xinyi Yin~\orcidlink{0009-0003-9553-8654}, Fei Yu~\orcidlink{0000-0003-0193-1790} \\ Wei
Zhou~\orcidlink{0000-0003-3641-1429}, Yanfei Zhong~\orcidlink{0000-0001-9446-5850}, Yang Liu~\orcidlink{0000-0002-1312-0146}, and Dingkang Yang~\orcidlink{0000-0003-1829-5671} 
\thanks{This research was supported by the Key R\&D Program of Zhejiang under Grant No. 2024C01036 and the Zhejiang Provincial Natural Science Foundation of China under Grant No. LQ23F030009.}

\thanks{Junxiao Xue and Kelu Yao are currently with the Research Center for Space Computing System, Zhejiang Lab, Hangzhou, 311100, China (E-mail: {\small xuejx@zhejianglab.cn});}
\thanks{Quan Deng is with the Hangzhou Institute for Advanced Study, University of Chinese Academy of Sciences, Hangzhou, 310024, China. (E-mail: {\small dengquan23@mails.ucas.ac.cn});}
\thanks{Xuecheng Wu is currently with the School of Computer Science and Technology, Xi'an Jiaotong University, Xi'an, 710049, China. (E-mail: {\small wuxc3@stu.xjtu.edu.cn});}
\thanks{Xinyi Yin is with the School of Cyber Science and Engineering, Zhengzhou University, Zhengzhou, 450002, China (E-mail: {\small yinxinyi@stu.zzu.edu.cn});}
\thanks{Fei Yu is with Liaoning University of Technology, Jinzhou, 123099, China (E-mail: {\small yufei\_hitcs@163.com});}
\thanks{Wei Zhou is with the School of Computer Science and Informatics, Cardiff University, Cardiff, CF24 4AG, United Kingdom. (E-mail: {\small zhouw26@cardiff.ac.uk});}
\thanks{Yanfei Zhong is with the State Key Laboratory of Information Engineering in Surveying, Mapping and Remote Sensing (LIESMARS), Wuhan University, Wuhan, 430072, China. (E-mail: {\small zhongyanfei@whu.edu.cn});}
\thanks{Yang Liu is with the College of Electronic and Information Engineering, Tongji University, Shanghai, 201804, China. (E-mails: yang\_liu@ieee.org).}
\thanks{Dingkang Yang is with the College of Intelligent Robotics and Advanced Manufacturing, Fudan University~\&~Fysics AI, Shanghai, 200433, China (E-mail: {\small dkyang20@fudan.edu.cn});}
\thanks{Junxiao Xue, Quan Deng, and Xuecheng Wu deserve equal contributions.}
\thanks{Work done during Quan Deng's research internship at Zhejiang Lab.}
\thanks{Corresponding authors: Dingkang Yang \& Yang Liu.}
}

\markboth{IEEE Transactions on Geoscience and Remote Sensing}%
{How to Use the IEEEtran \LaTeX \ Templates}

\maketitle

\begin{abstract}
Remote sensing change understanding (RSCU) is essential for analyzing remote sensing images and understanding how human activities affect the environment. However, existing datasets lack deep understanding and interactions in the diverse change captioning, counting, and localization tasks. To tackle these gaps, we construct ChangeIMTI, a new large-scale interactive multi-task instruction dataset that encompasses four complementary tasks including change captioning, binary change classification, change counting, and change localization. Building upon this new dataset, we further design a novel vision-guided vision-language model (ChangeVG) with dual-granularity awareness for bi-temporal remote sensing images (\textit{i.e.}, two remote sensing images of the same area at different times). The introduced vision-guided module is a dual-branch architecture that synergistically combines fine-grained spatial feature extraction with high-level semantic summarization. These enriched representations further serve as the auxiliary prompts to guide large vision-language models (VLMs) (\textit{e.g.}, Qwen2.5-VL-7B) during instruction tuning, thereby facilitating the hierarchical cross-modal learning. We extensively conduct experiments across four tasks to demonstrate the superiority of our approach. Remarkably, on the change captioning task, our method outperforms the strongest method Semantic-CC by 1.39 points on the comprehensive $S^*_m$ metric, which integrates the semantic similarity and descriptive accuracy to provide an overall evaluation of change caption. Moreover, we also perform a series of ablation studies to examine the critical components of our method. The source code and associated data for this work are publicly available at~\href{https://github.com/Quan-zzx/ChangeVG}{Github}.
\end{abstract}


\section{Introduction}
\IEEEPARstart{I}{n} recent years, remote sensing change understanding (RSCU) techniques have emerged as powerful tools for monitoring land cover dynamics~\cite{land_cover}, urban expansion~\cite{urban_expansion}, environmental degradation~\cite{env_detect}, and disaster management~\cite{disasters}. A key focus of these techniques has been change detection, which aims to identify differences between bi-temporal (captured at two different time points) remote sensing images.

Despite significant progress in remote sensing change detection in recent years, particularly in pixel-level and region-level analysis using widely adopted deep learning architectures such as Siamese networks, encoder-decoder frameworks, and semantic segmentation models~\cite{change_survey,change_survey1,change_survey2}, existing methods remain primarily focused on identifying the spatial location and extent of changes. However, they are generally unable to generate contextual or semantic descriptions of the detected changes effectively~\cite{change_agent,change_chat}. Moreover, these approaches generally provide limited support for interactive exploration, making it difficult for users to query or interpret changes in a flexible, task-driven manner. Consequently, there is a growing need to move beyond localization-oriented change detection toward comprehensive change understanding and natural language captioning. Describing changes through natural language can facilitate human-understandable interpretation and enable higher-level reasoning in remote sensing analysis.

With the rapid developments of VLMs, these models have demonstrated exceptional capabilities in joint vision-language understanding and reasoning. They have been widely applied in various tasks involving natural images, such as image captioning and visual question answering (VQA)~\cite{mllm_survey,mllm_survey_1}. Due to their ability to align visual information with natural language representations effectively, VLMs are emerging as promising tools for addressing semantic understanding challenges in remote sensing images. Recent studies have begun to explore the use of VLMs for change captioning in remote sensing, generating natural language descriptions of differences between image pairs. However, most efforts remain limited to static caption generation and have not fully leveraged the interactive and multi-task reasoning capabilities of VLMs~\cite{Semantic-CC}. Such as change counting, spatial localization, and comprehensive captioning remain insufficiently addressed. Therefore, a more unified and comprehensive approach is  required, one that moves beyond simple captioning and supports interactive, question driven understanding of changes in the remote sensing images.

In this work, we construct a new large-scale multi-task instruction-tuning dataset for remote sensing change understanding, named ChangeIMTI. The dataset not only includes the task of change captioning, but also extends to a variety of VQA tasks, such as determining whether a change has occurred, counting the number of changed objects, and localizing the regions of change. These tasks are designed to comprehensively enhance the model's semantic understanding and interactive reasoning capabilities in the context of remote sensing images. By jointly training on these complementary tasks, the model is able to develop more robust and generalizable multimodal representations. For instance, binary change classification improves the model’s awareness of changes, while change counting enhances sensitivity to object-level differences. The resulting multi-task synergy not only improves individual task performance but also leads to better overall generalization across diverse remote sensing scenarios.

With this dataset, we propose a unified framework that incorporates a vision-guided  module to enhance fine-grained spatial and semantic feature extraction. After receiving bi-temporal remote sensing image pairs, the vision-guided module processes them through two parallel branches: one branch is designed to extract fine-grained visual cues such as spatial location and object count, which serve as detailed prompts to improve the model’s sensitivity to subtle changes; the other branch focuses on capturing coarse-grained semantic information, such as overall scene-level differences relevant to change captioning. These visual features are subsequently integrated into a VLM, which is fine-tuned on the ChangeIMTI. This design facilitates more interpretable and interactive remote sensing change understanding by tightly coupling detailed visual perception with high-level semantic reasoning.

In conclusion, the main contributions of this paper can be summarized as follows:

\begin{itemize} 
\item[$\bullet$] We construct a new large-scale multi-task instruction-tuning dataset denoted ChangeIMTI for RSCU. It covers change captioning, binary change classification, change counting, and change localization tasks, providing a unified testbed for applying VLMs in the remote sensing change domain.

\item[$\bullet$] 
We propose a unified VLM-based framework for remote sensing change understanding, equipped with a vision-guided module that achieves dual-granularity change perception through simultaneous modeling of fine-grained and global representations. The framework supports both change captioning and VQA, thereby enhancing semantic interpretability and interactive reasoning.

\item[$\bullet$] We have conducted extensive evaluations across multiple downstream remote sensing change understanding tasks and achieve impressive model performance, demonstrating the effectiveness and generalizability of our introduced approach.

\end{itemize}

\section{Related Work}

\subsection{Remote Sensing Change Datasets}
Remote sensing change datasets can generally be categorized into three types based on the form of change annotation they provide, each serving different research purposes and application scenarios. The first type includes only binary change masks (\textit{e.g.}, LEVIR-CD~\cite{LEVIR-CD}, WHU-CD~\cite{WHU-CD}), which are primarily designed for pixel-level change detection, offering high-resolution satellite or aerial images annotated with binary labels indicating whether a change has occurred. However, these datasets typically lack semantic information about the nature or category of the change, limiting their utility for higher-level interpretation or reasoning tasks. The second type focuses on textual descriptions of scene changes, without providing explicit pixel-level masks. Datasets such as RSICCFormer~\cite{RSICCFormer} include natural language annotations that describe how a scene has changed over time, capturing semantic and contextual information. However, they lack fine-grained mask annotations, limiting their effectiveness in tasks that require precise spatial localization of changes. The third type offers both binary change masks and corresponding textual descriptions, enabling multi-level and multi-modal change understanding. A representative example is LEVIR-MCI~\cite{change_agent}. Although the dataset incorporates both fine-grained spatial annotations and high-level semantic descriptions, it was not specifically designed or optimized for use with VLMs, limiting its applicability in this emerging research area.

\subsection{Remote Sensing Change Understanding}
RSCU requires not only generating captions for change images but also capturing fine-grained differences. A key subtask, Remote Sensing Change Captioning (RSCC), aims to generate natural language descriptions of semantic differences between bi-temporal remote sensing images~\cite{change_agent}. Early methods followed image captioning pipelines with CNN encoders and RNN/Transformer decoders, often introducing fusion strategies to highlight change regions~\cite{caption3, caption2, caption1, change_caption, chouaf2021captioning}. but struggled with subtle variations. More recently, the emergence of VLMs has introduced new paradigms for addressing RSCC tasks. The approach leverages powerful pretrained models such as CLIP for visual representation and GPT-style LLMs for text generation, enabling improved semantic reasoning and generalization, such as Liu \textit{et al.}~\cite{liu2023decoupling}. However, such two-stage paradigms remain relatively underexplored. Moreover, despite these advancements, existing RSCC models continue to face significant challenges, including the accurate description of small or subtle changes, fine-grained spatial localization, and effective contextual reasoning, particularly in complex or cluttered remote sensing.

In summary, although RSCC has achieved notable progress through the use of deep learning techniques and the availability of curated benchmark datasets, existing approaches remain limited in their ability to deliver human-aligned and semantically rich descriptions. And current research still lacks sufficient exploration into interactive methods for fine-grained change understanding, leaving a critical gap in fully leveraging these models for detailed change analysis. The integration of advanced VLMs presents a promising avenue for addressing these limitations. These models offer the potential to enhance contextual reasoning, align fine-grained visual changes with language outputs, and support more interactive and interpretable remote sensing applications.

\subsection{Vision-language Models in Remote Sensing}
VLMs have demonstrated strong potential in enhancing image understanding by integrating visual and linguistic information~\cite{qwen2.5-vl,zhang2025tokenfocus,Hkd4vlm}, particularly in remote sensing~\cite{Remoteclip, Rsgpt, Geochat}. Current studies mainly follow two directions: (1) pretraining on large-scale remote sensing image datasets using self-supervised learning, and (2) fine-tuning existing VLMs with small but high-quality remote sensing datasets.

For pretraining, RingMo~\cite{RingMo} applied masked image modeling on 2 million remote sensing images to reduce the domain gap with natural images. Other efforts explored CNN–ViT combinations~\cite{wang2022advancing} and extended to SAR or spatio-temporal modalities. RemoteCLIP~\cite{Remoteclip} pioneered vision-language alignment in this domain, achieving strong zero-shot transfer on 16 tasks. GRAFT~\cite{mall2023remote} aligned satellite and ground images without textual supervision. On the fine-tuning side, RSGPT~\cite{Rsgpt} instruction-tuned InstructBLIP with curated caption datasets, while RSPrompter~\cite{RSPrompter} and TTP~\cite{TTP} enhanced SAM for remote sensing instance segmentation and change detection.

Despite progress, remote sensing VLMs still face several challenges. Compared to natural images, remote sensing data exhibits greater scale variability, more complex spatial-temporal structures, and sparse semantic cues, making cross-modal alignment particularly difficult, while high computational costs limit deployment. Future research should explore lightweight fine-tuning, self-supervised cross-modal alignment, and domain adaptation to enhance the generalizability, interactivity, and reasoning capacity of VLMs in remote sensing applications.
\section{ChangeIMTI}
\label{chap:ChangeIMTI}

In this section, we first provide a detailed description of the constructed ChangeIMTI dataset, including task formulation, data sources, annotation format, and instruction generation strategy. Subsequently, we provide a detailed description of the construction process for each category of task data. Finally, we present the statistical characteristics of the dataset.

\begin{figure}[!t]
    \centering
    \includegraphics[width=\linewidth]{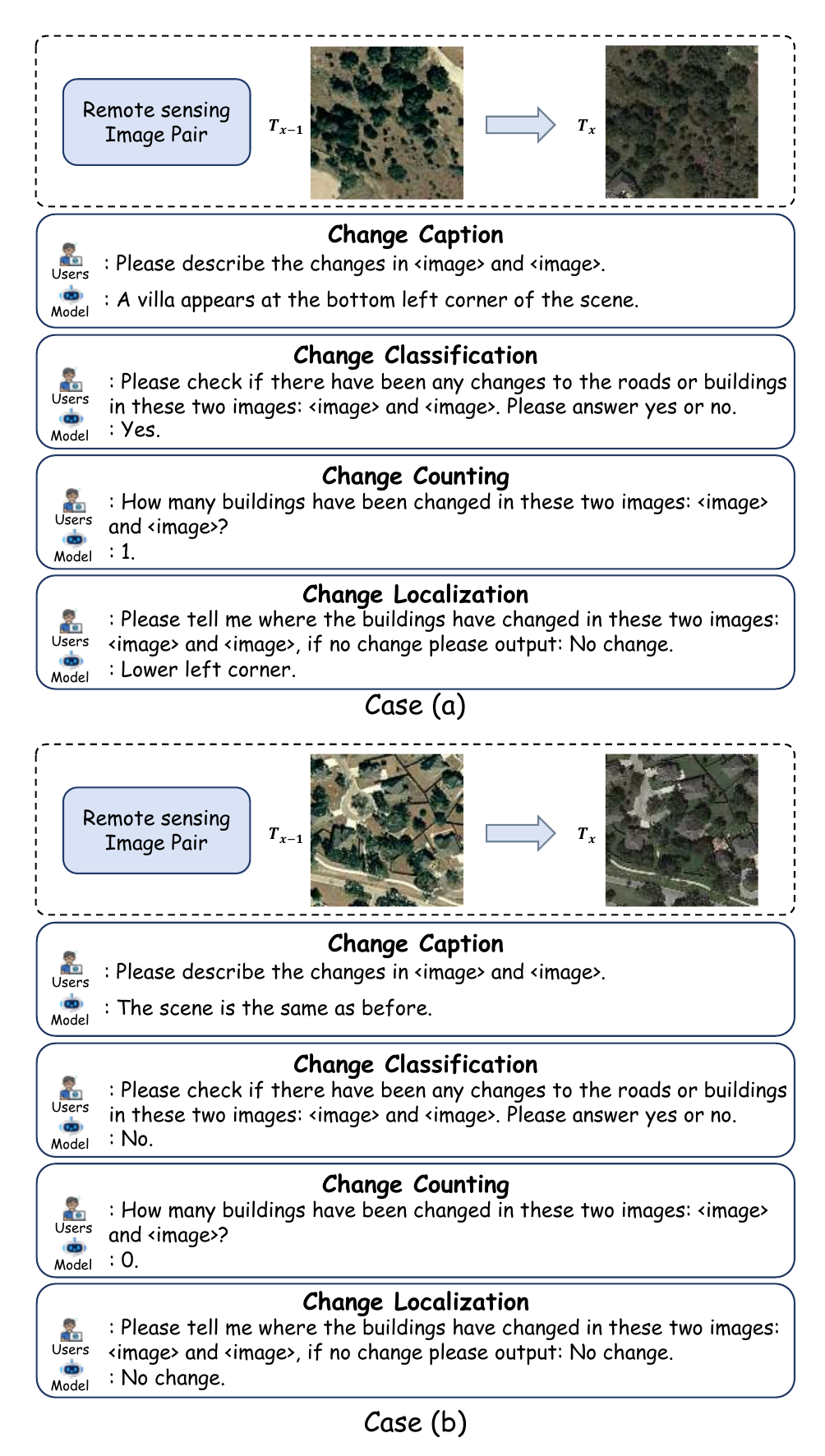}
    \caption{Examples from the ChangeIMTI dataset illustrating two scenarios. The $T_{x-1}$ and $T_x$ represent images from two different moments in time, with $T_{x-1}$ being the image from the earlier time and $T_x$ being the one from the later time. In Case (a), the image pair contains visible changes. In Case (b), the image pair remains unchanged. The four subtasks (change captioning, binary change classification, change counting, and change localization) demonstrate how changes are described, detected, quantified, and localized, respectively.}
    \label{fig:show_data}
\end{figure}

LEVIR-CC~\cite{RSICCFormer} represents a large-scale dataset specifically designed for remote sensing change captioning tasks, containing 10,077 pairs of 256×256 satellite images collected from 20 regions in Texas, USA, with temporal spans ranging from 5 to 15 years. The dataset maintains a balanced composition with 5,038 pairs depicting actual land-cover changes and 5,039 pairs showing no changes, each meticulously annotated with five reference captions to capture description variability. On top of this dataset, LEVIR-MCI~\cite{change_agent} provides pixel-level change annotations, making it suitable for change detection tasks. While these datasets are valuable for basic change captioning and detection, they are insufficient to fully exploit the capabilities of VLMs, especially in terms of fine-grained semantic reasoning and diverse VQA. To address this limitation, we construct a large-scale, multi-task instruction-tuning dataset based on LEVIR-CC and LEVIR-MCI. The proposed dataset supports a range of tasks including change captioning, binary change classification, change counting, and change localization, aiming to enhance the generalization and interaction capabilities of VLMs in RSCU. As shown in Fig.~\ref{fig:show_data}, an example from our dataset illustrates task performance under changed and unchanged conditions.

\begin{table}[!t]
    \centering
    \caption{Statistics of the ChangeIMTI instruction tuning dataset, detailing the number of instances across different task types.}
    \label{tab:ChangeIMTI}
    \begin{tabular}{p{3.5cm} P{3.5cm}}
        \toprule
         \textbf{Task Type} &  \textbf{Data Number}    \\
        \midrule
        Change Caption             &  \num{24444}  \\
        Binary Change Classification      &  \num{8148}   \\
        Change Counting            &  \num{16296}  \\
        Change Localization        &  \num{16296}  \\
        Multi-turn Conversation    &  \num{24444}  \\
        \textbf{Total}                      &  \num{89628} \\
        \hline
    \end{tabular}
\end{table}

\subsection{Change Captioning}
Change captioning as the core task, requiring the model to generate natural language descriptions of differences between bi-temporal remote sensing images. To enhance linguistic diversity and improve generalization, we construct five instruction–response pairs for each image pair using the five reference captions from LEVIR-CC. Each instruction is framed to encourage the VLMs to interpret and articulate the semantic differences between the two images. 

The task emphasizes the model’s ability to understand complex scene changes (\textit{e.g.}, new buildings, demolished roads) and translate them in natural language. Specifically, the user’s instruction requires the model to describe the change. The model’s output should be a natural-language narrative that delineates this change in detail, including the specific category of the change, the number and locations of the affected elements, and any other pertinent characteristics relevant to the revision. For unchanged image pairs, we construct a single instruction where the expected model response is a negation, such as “the scene is the same as before.” helping prevent overfitting toward always generating change descriptions. One example is displayed as follows:
\begin{findingsbox}
\item Instruction Template: \\
    \hspace*{1em} User: ``Please describe the changes in $<$image$>$ and $<$image$>$.'' \\
    \hspace*{1em} VLM: [Natural language caption]
\end{findingsbox}

\subsection{Binary Change Classification}
To introduce decision-making capability, we formulate a binary classification task that determines whether a change has occurred between two images. The task simplifies the complex problem of change detection into a straightforward yes/no decision, making it more manageable for computational models. Specifically, the user’s input is a query asking whether any changes have occurred, and the model is expected to respond with only “yes” or “no.”
\begin{findingsbox}
\item Instruction Template: \\
    \hspace*{1em} User: ``Please check if there have been any changes to the roads or buildings in these two images:$<$image$>$ and $<$image$>$. Please answer yes or no.'' \\ 
    \hspace*{1em} VLM: ``Yes'' / ``No''
\end{findingsbox}

\subsection{Change Counting}
Change counting introduces object-level granularity by requiring the model to estimate the number of discrete change regions. To obtain the reference counts, we preprocess the binary change masks using OpenCV’s contour detection algorithm, which identifies and counts connected components (\textit{i.e.}, contiguous areas of change). The process ensures that the detected changes are precisely segmented and counted. The task encourages the VLMs to develop instance awareness of changes, going beyond binary classification to quantify scene modifications and identify the specific number of altered regions, thereby enhancing the model’s ability to analyze complex scene dynamics. Specifically, the user’s query will ask for the number of changes to roads or buildings, and the model’s response should be an Arabic numeral providing that count.
\begin{findingsbox}
\item Instruction Template: \\
    \hspace*{1em} User: ``How many roads[or buildings] have been changed in these two images:$<$image$>$ and $<$image$>$?''\\
    \hspace*{1em} VLM: [Number of detected change regions]
\end{findingsbox}

\subsection{Change Localization}
To evaluate spatial reasoning, we construct a change localization task that maps each identified change to a coarse-grained spatial region. The task helps the model better understand spatial context by associating changes with specific areas in the image. Specifically, we compute the centroid of each change mask and map it to one of nine predefined regions in a 3×3 grid: top left corner, left, lower left corner, top, center, lower, top right corner, right, and lower right corner. For each image pair, we generate instructions querying the location of changes in specific semantic categories, such as buildings or roads. Specifically, when the user asks about the locations of changes to roads or buildings, the model should respond with one or more directional indicators. The localization process provides additional spatial context, allowing the model to not only detect changes but also infer their positions within the overall scene. The fine-grained localization guides the model's spatial reasoning abilities.
\begin{findingsbox}
\item Instruction Template: \\
    \hspace*{1em} User: ``Please tell me where the roads[or buildings] have changed in these two images:$<$image$>$ and $<$image$>$, if no change please output: No change.''\\
    \hspace*{1em} VLM: [top left corner, left, lower left corner, top, center, lower, top right corner, right, lower right corner](Select a few of them)
\end{findingsbox}

\subsection{Multi-turn Dialogue}
To simulate real-world user interactions, we synthesize multi-turn conversations that sequentially and seamlessly combine the above tasks into a coherent dialogue. The design follows a structured progression: starting with general change inquiries, the conversations incrementally deepen into specific queries about captioning, classification, counting, and location of changes. This hierarchical approach begins with basic questions, progresses to quantitative analysis, locational identification, and culminates in descriptive analysis. To further align with the stochastic nature of human inquiry patterns, we introduce variability by integrating multiple tasks within a single dialogue in a randomized manner. This dual strategy of combining systematic task progression with randomized task interleaving significantly enhances the model's multi-step reasoning ability, robustness to instruction order variations, and consistency in longer conversational contexts. Such synthetic dialogues not only strengthen the model's analytical capabilities but also lay the groundwork for interactive AI agents in various remote sensing analytics.

\subsection{The Statistics of ChangeIMTI}
In summary, our constructed dataset comprises five components, \textit{i.e.}, change captioning, binary change classification, change counting, change localization, as well as multi-turn conversation. The change caption task provides textual descriptions of the detected changes, while binary change classification determines whether changes have occurred. change counting and change localization offer fine-grained information about the changes, and multi-turn dialogue enables interactive capabilities. As displayed in Table~\ref{tab:ChangeIMTI} above, it totally contains 89,628 data samples, making it the currently largest instruction-tuning dataset for change detection tasks.

\section{Methodology}

\begin{figure*}[!h]
    \centering
    \includegraphics[width=\linewidth]{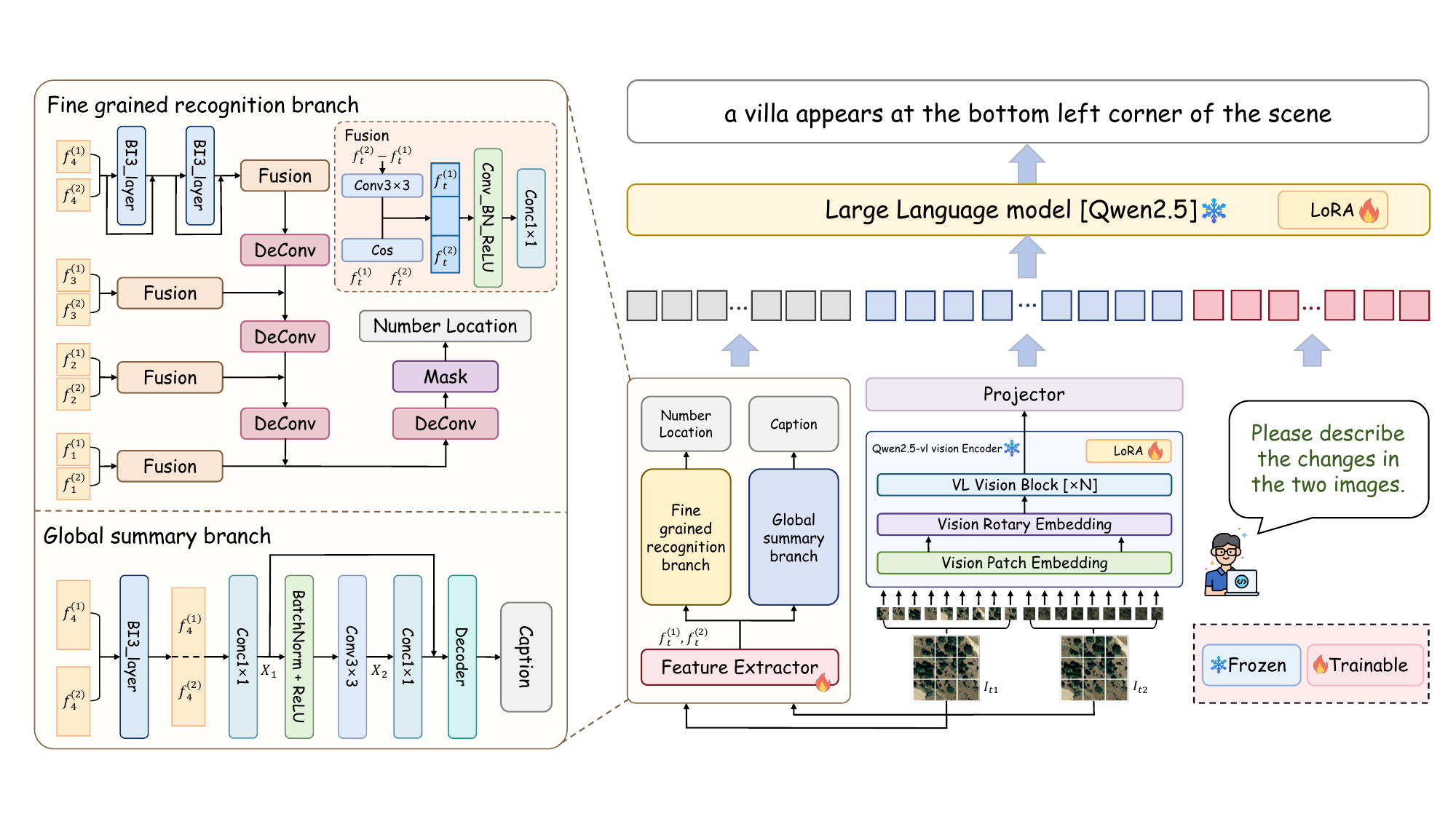}
    \caption{Overview of ChangeVG. The right panel illustrates the overall architecture of our model, which takes the input image and the user query, processes them through Qwen2.5-VL-7B and a vision-guided module, and finally generates a response to the query. The left panel presents the details of the vision-guided module, which consists of a global summary branch and a fine-grained recognition branch. These branches extract coarse-grained caption information and fine-grained details, respectively.}
    \label{fig:framework}
\end{figure*}

As illustrated in Fig.~\ref{fig:framework}, our model consists of two main components: a novel vision-guided module and a VLM backbone. The vision-guided module is designed to extract auxiliary visual cues with dual-granularity semantics, capturing both fine-grained details such as object count and spatial locations and holistic scene-level descriptions in caption format. In contrast to the visual features extracted by ViT, which primarily encode basic image patterns, our module delivers richer and more semantically meaningful representations. These dual-granularity signals are then integrated into the VLM, which jointly processes the bi-temporal remote sensing images and the injected prompts to generate context-aware responses.

\subsection{Vision-guided Module}

To enhance the instruction-following capability of our VLMs in remote sensing change captioning, we introduce a vision-guided module that processes bi-temporal remote sensing images to extract dual-granularity visual cues, encompassing both fine-grained details and global-level semantics. As shown in Fig.~\ref{fig:framework}, the module comprises three components: a feature extractor, a fine-grained recognition branch, and a global summary branch.

\subsubsection{Feature Extractor}
The feature extraction module is designed to capture rich, multi-scale semantic representations from bi-temporal remote sensing images, serving as the visual backbone for both the fine-grained recognition branch and the global summary branch. Specifically, the two input images acquired at different time points are independently processed by identical encoder networks with shared parameters, based on the SegFormer-B1 encoders~\cite{segformer}. The parameter-sharing strategy ensures consistent feature alignment across temporal dimensions while producing hierarchical visual features at multiple levels of abstraction.

To formally denote the extracted features, let $I_{t1}$ and $I_{t2}$ represent the input images at two different time points. The shared encoder $\mathcal{F}_\theta$ (SegFormer-B1) extracts a set of multi-level features from each image:
\begin{equation}
\{f_1^{(1)}, f_2^{(1)}, f_3^{(1)}, f_4^{(1)}\} = \mathcal{F}_\theta(I_{t1}), 
\end{equation}
\begin{equation}
\{f_1^{(2)}, f_2^{(2)}, f_3^{(2)}, f_4^{(2)}\} = \mathcal{F}_\theta(I_{t2}),
\end{equation}
where $f_l^{(i)}$ denotes the feature map at level $l \in \{1,2,3,4\}$ extracted from image $I_{ti}$. These features encode spatial and semantic cues at increasing levels of abstraction.

The extracted features from both images are subsequently forwarded to the downstream fine-grained recognition branch and global summary branch, which are responsible for detailed instance reasoning and caption generation, respectively.

\subsubsection{Fine-Grained Recognition Branch}
The fine-grained recognition branch enhances the model's capacity to detect and interpret localized changes between bi-temporal remote sensing images. Based on multi-level features extracted by the shared SegFormer-B1 encoders, this branch performs progressive fusion and decoding to generate a high-resolution change mask and fine-grained change attributes.

The multi-scale feature maps from time $t_1$ and $t_2$ be represented as:
\begin{equation}
f_t^{(1)} = \left\{ f_1^{(1)}, f_2^{(1)}, f_3^{(1)}, f_4^{(1)} \right\},
\end{equation}
\begin{equation}
f_t^{(2)} = \left\{ f_1^{(2)}, f_2^{(2)}, f_3^{(2)}, f_4^{(2)} \right\}.
\end{equation}

For each level $i$, a similarity-guided fusion module combines features using cosine similarity and convolution:
\begin{align}
\alpha_i &= \cos(f_i^{(1)}, f_i^{(2)})+\text{Conv}_{3 \times 3}(f_i^{(2)}-f_i^{(1)}) , \\
\tilde{f}_i &= \text{Conv}_{3 \times 3}\left(\text{Concat}[f_i^{(1)} , \alpha_i , f_i^{(2)}]\right), \\
f_i^{\text{fused}} &= \text{Conv}_{1 \times 1}\left(\text{ReLU}\left(\text{BN}(\tilde{f}_i)\right)\right).
\end{align}
The top-level fused features $f_4^{\text{fused}}$ are further processed by a series of stacked BI3 layers~\cite{change_agent} to enhance their semantic representation, which can be formulated as:

\begin{equation}
f_4^{\mathrm{refined}} = \mathrm{BI3Layer}_2\left( \mathrm{BI3Layer}_1\left( f_4^{\mathrm{fused}} \right) \right).
\end{equation}

Next, fused features from all levels are progressively decoded through deconvolution layers. The decoding process can be represented as:
\begin{align}
d_4 &= \text{DeConv}(f_4^{\text{refined}}), \\
d_i &= \text{DeConv}(f_i^{\text{fused}} \oplus d_{i+1}), \quad \text{for } i = 3, 2, 1,
\end{align}
where $\oplus$ denotes channel-wise concatenation.

The final decoded feature map $d_1$ is passed through a $1 \times 1$ convolution with sigmoid activation to produce the binary change mask:
\begin{equation}
M = \sigma\left( \mathrm{Conv}_{1 \times 1}(d_1) \right).
\end{equation}
We utilize OpenCV-based post-processing techniques to analyze the predicted binary change mask $M$. Through contour detection and connected component analysis, we extract the number and spatial locations of changed instances, specifically focusing on road and building changes. These attributes serve as auxiliary cues for downstream reasoning tasks such as change counting and localization.

\subsubsection{Global Summary Branch}
The global summary branch is designed to provide an initial semantic understanding of the scene-level changes between bi-temporal remote sensing images. In contrast to the fine-grained recognition branch, which focuses on localized and detailed change patterns, this branch generates a coarse-grained summary that serves as a visual prompt to guide the subsequent reasoning process of the VLMs. By capturing the overall transformation trends in the input image pair, the global summary branch offers high-level contextual information that facilitates more coherent and accurate language generation in downstream tasks such as change captioning.

Specifically, we feed the high-level semantic features $f_4^{(1)}$ and $f_4^{(2)}$, extracted from the two temporal images, into the BI3 module to enhance and fuse the global contextual representations across time. The resulting bi-temporal feature is then passed through a convolutional transform layer, which transforms the visual features from the image domain into a semantic embedding space suitable for language modeling. Finally, these embeddings are decoded by a Transformer-based language decoder to produce an initial coarse-grained change description, which serves as a prompt for subsequent reasoning by the VLMs. The transform layer $Trans(\cdot)$ can be formally represented as follows:
\begin{align}
\mathrm{Trans}(f^{(1)},f^{(2)}) &= X_1 + \mathrm{conv}_{1\times1}(X_2),
\label{eq:trans_func} \\
X_1 &= \mathrm{conv}_{1\times1}\left(\mathrm{concat}\left([f^{(1)},f^{(2)}]\right)\right) ,\label{eq:x1} \\
X_2 &= \mathrm{ConvBNReLU}_{3\times3}\left(\mathrm{conv}_{1\times1}(X_1)\right) .\label{eq:x2}
\end{align}
The transformed representation is then fed into a Transformer-based language decoder to produce a coarse-grained natural language description of the detected changes. The caption summarizes the global transformation context and serves as a visual prompt to guide subsequent fine-grained reasoning in the downstream VLMs.

\subsection{Instruction Tuning for ChangeVG}
To enable the model to effectively understand and respond to queries based on bi-temporal remote sensing images, we fine-tune the open-source VLMs Qwen2.5-VL-7B~\cite{qwen2.5-vl} on our constructed \textit{ChangeIMTI} dataset. Qwen2.5-VL-7B is a vision-language model built upon the Qwen2.5-7B language backbone, equipped with a visual encoder for image understanding and a multi-layer cross-modal fusion module to align image and text representations.

Our instruction tuning aligns the model's generative capabilities with multi-task RSCU. Each training sample is formatted as an instruction–response pair:
\begin{equation}
\mathcal{D} = \{(I_{t1}, I_{t2}, q, r)\},
\end{equation}
where $I_{t1}$ and $I_{t2}$ represent the bi-temporal remote sensing images, $q$ is the user query (\textit{e.g.}, “Please describe the changes.”), and $r$ is the target response generated by human annotators.

\begin{figure}[t!]
\centering
\includegraphics[width=\linewidth]{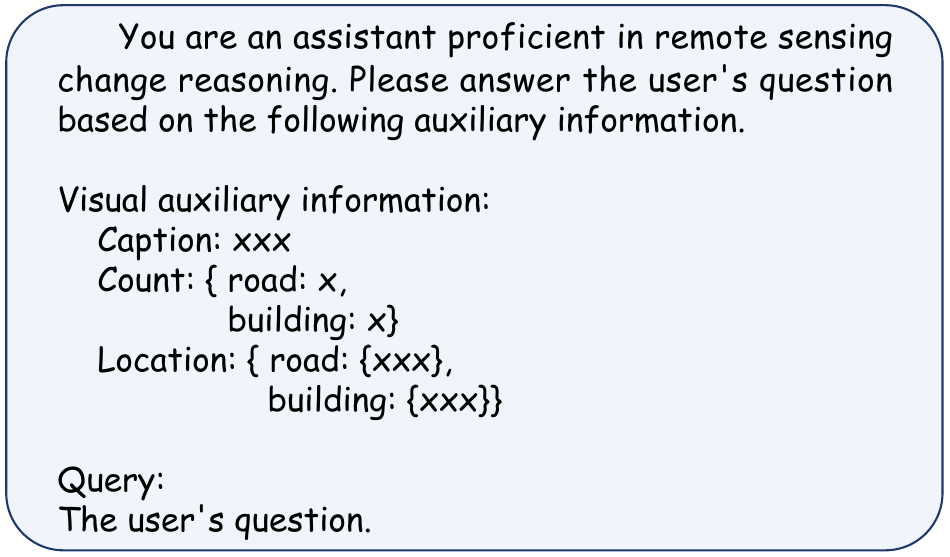}
\caption{The example of our prompt. It illustrates the architecture of the final prompt fed into the VLM.}
\vspace{-0.5em}
\label{fig:prompt}
\end{figure}

To better exploit both fine-grained and global summary information, we design a prompt. Specifically, we leverage a vision-guided module to extract coarse-grained scene-level captions and fine-grained change indicators, including object counts and spatial localization cues. These outputs are translated into natural language and injected into the final prompt as structured contextual priors. The fusion module then integrates these visual priors with the user query and corresponding image pair to construct a comprehensive multimodal prompt display in Fig.~\ref{fig:prompt}. The enriched prompt serves as input to the instruction-tuned LLM, enabling it to reason over both high-level semantics and low-level details during generation. By explicitly encoding the auxiliary visual cues alongside user intent, the fusion mechanism significantly enhances the model’s capacity for precise, coherent, and contextually grounded change reasoning. Moreover, in multi-turn dialogues, we merge historical conversation with subsequent prompts to support interactive user engagement.

\section{Experiments}

\begin{table*}[!t]
    \centering
    \caption{Compare with SOTA methods on the change caption task. The best results are highlighted in bold. The second-best results are indicated with underlines.}
    \label{tab:change caption}
    \begin{tabular}{p{3.5cm} | P{2.2cm} P{2.2cm} P{2.2cm} P{2.2cm} | P{2.2cm}}
        \toprule
         \textbf{Method} &  \textbf{BLEU-4} & \textbf{METEOR} & \textbf{ROUGE$_L$} & \textbf{CIDEr-D} & \textbf{$S_m^{*}$} \\
        \midrule
        DUDA ~\cite{DUDA}                        & 57.79 & 37.15 & 71.04 & 124.32 & 72.58 \\
        MCCFormer-S ~\cite{MCCFormer}            & 56.68 & 36.17 & 69.46 & 120.39 & 70.68 \\
        MCCFormer-D ~\cite{MCCFormer}            & 56.38 & 37.29 & 70.32 & 124.44 & 72.11 \\
        RSICCFormer \cite{RSICCFormer}           & 62.77 & 39.61 & 74.12 & 134.12 & 77.66 \\
        PSNet   \cite{PSNet}                     & 62.11 & 38.80 & 73.60 & 132.62 & 76.78 \\
        Chg2Cap  \cite{Chg2Cap}                  & 64.39 & 40.03 & 75.12 & 136.61 & 79.04 \\
        Prompt-CC  \cite{liu2023decoupling}      & 63.54 & 38.82 & 73.72 & 136.44 & 78.13 \\
        Semantic-CC \cite{Semantic-CC}           & \underline{64.51} & \underline{40.58} & \textbf{77.76} & \underline{138.51} & \underline{80.34} \\
        \midrule
        ChangeVG (Ours)                         & \textbf{65.08} & \textbf{42.06} & \underline{76.95} & \textbf{142.83} & \textbf{81.73} \\
        \bottomrule
    \end{tabular}
\end{table*}

\subsection{Experimental settings}
In this work, all the training and evaluations of our introduced model are conducted on a machine with 8 $\times$ NVIDIA 4090 GPUs. For comparisons with state-of-the-art VLMs, we deploy their corresponding APIs. The test set employed in our experiments is derived from the LEVIR-MCI dataset, following the preprocessing procedures described in Section ~\ref{chap:ChangeIMTI}, which can ensure the consistency between the training and evaluation data format.

\subsection{Evaluation Metrics}
To evaluate both the effectiveness and the generalizability of our proposed approach, we conduct experiments not only on the change captioning task but also on three additional tasks: binary change classification, change counting, and change localization.
\begin{itemize} 
    \item Change captioning task aims to describe the differences between two images captured at different times using natural language.
    \item Binary change classification task aims to determine whether a change has occurred in a specific object or region between two images captured at different times.
    \item Change counting task aims to estimate the number of changed objects of a specified category between two images taken at different time points.
    \item Change localization task aims to identify the spatial locations of changed objects within the image pair, indicating where changes have occurred over time.
\end{itemize}

For the change captioning task, we evaluate the model using BLEU-4~\cite{bleu}, METEOR~\cite{meteor}, ROUGE-L~\cite{rouge}, and CIDEr-D~\cite{cider} to assess 4-gram precision, alignment with human judgment (considering synonymy, stemming, and word order), the ability to capture the longest common subsequence, and consensus with human-written descriptions based on TF-IDF weighting of n-grams, respectively.
To comprehensively evaluate the model’s performance on the change captioning task, we use an aggregate metric $S_m^{*}$ ~\cite{s_star}, which is defined as the average of four widely-used evaluation metrics: BLEU-4, METEOR, ROUGE-L, and CIDEr-D. The equation is as follows:
\begin{equation}
\scalebox{0.85}{$
S_m^* = \frac{1}{4} \times \big( BLEU_4 + METEOR + ROUGE_L + CIDEr - D ).$}
\end{equation}
The metric provides a balanced assessment by integrating lexical precision, semantic similarity, and n-gram consensus between the generated and reference captions.

For the binary change classification task, we evaluate the model using four standard metrics: Accuracy, Precision, Recall, and F1-score. These metrics comprehensively reflect the model’s classification performance by measuring its overall correctness (Accuracy), its ability to avoid false positives (Precision), its sensitivity to actual changes (Recall), and the balance between Precision and Recall (F1-score).

For the change counting task, we evaluate the model performance using Mean Absolute Error (MAE), applied separately to the road and building categories. MAE quantifies the average absolute difference between the predicted and ground-truth counts, providing an intuitive measure of the model’s counting accuracy for each type of changed object.

For the change localization task, we evaluate the model performance on both road and building categories using example-based accuracy, micro precision, micro recall, micro F1-score, and subset accuracy. These metrics jointly assess the model’s ability to accurately and completely localize changes across multiple categories.

\subsection{Main Results}

\subsubsection{Change Captioning}
We compare our method with multiple SOTA approaches, including DUDA~\cite{DUDA}, MCCFormer-S, MCCFormer-D \cite{MCCFormer}, RSICCFormer~\cite{RSICCFormer}, PSNet~\cite{PSNet}, Chg2Cap~\cite{Chg2Cap}, Prompt-CC~\cite{liu2023decoupling}, Semantic-CC~\cite{Semantic-CC}.

As shown in table~\ref{tab:change caption}, our model achieves the highest performance across BLEU-4, METEOR, and CIDEr-D metrics, with relative improvements of 1.07\%, 3.65\%, and 3.12\% over the best existing method, respectively. In addition, our approach achieves the best overall score on the aggregated metric $S_m^*$, outperforming Semantic-CC by 1.39 points. These results demonstrate the advantage of integrating fine-grained and global visual cues into the VLMs, leading to more accurate and contextually rich change descriptions.

\begin{figure*}[!t]
    \centering
    \includegraphics[width=\linewidth]{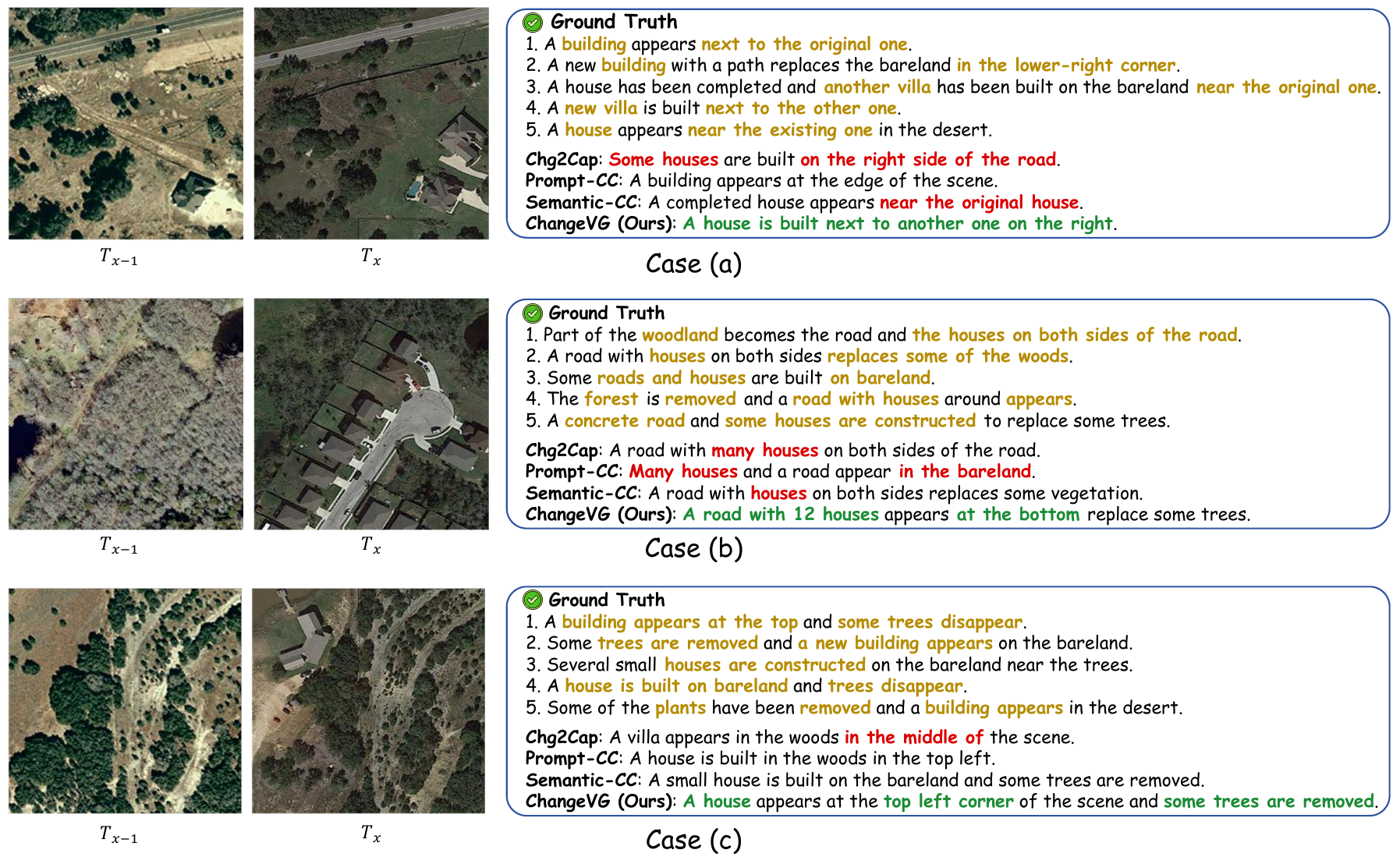}
    \caption{A direct comparison of our method against existing approaches on the change captioning task, where green indicates correct descriptions and red highlights inaccurate or inappropriate ones.}
    \label{fig:show_caption}
\end{figure*}

To more intuitively assess the effectiveness of our proposed method, Fig.~\ref{fig:show_caption} presents a visual comparison using three representative image samples, highlighting the differences between our approach and existing methods. As observed in the figure, our method outperforms other approaches in both quantity and location recognition, accurately identifying the number of changed objects as well as their spatial positions.

\begin{figure}[!t]
    \centering
    \includegraphics[width=\linewidth]{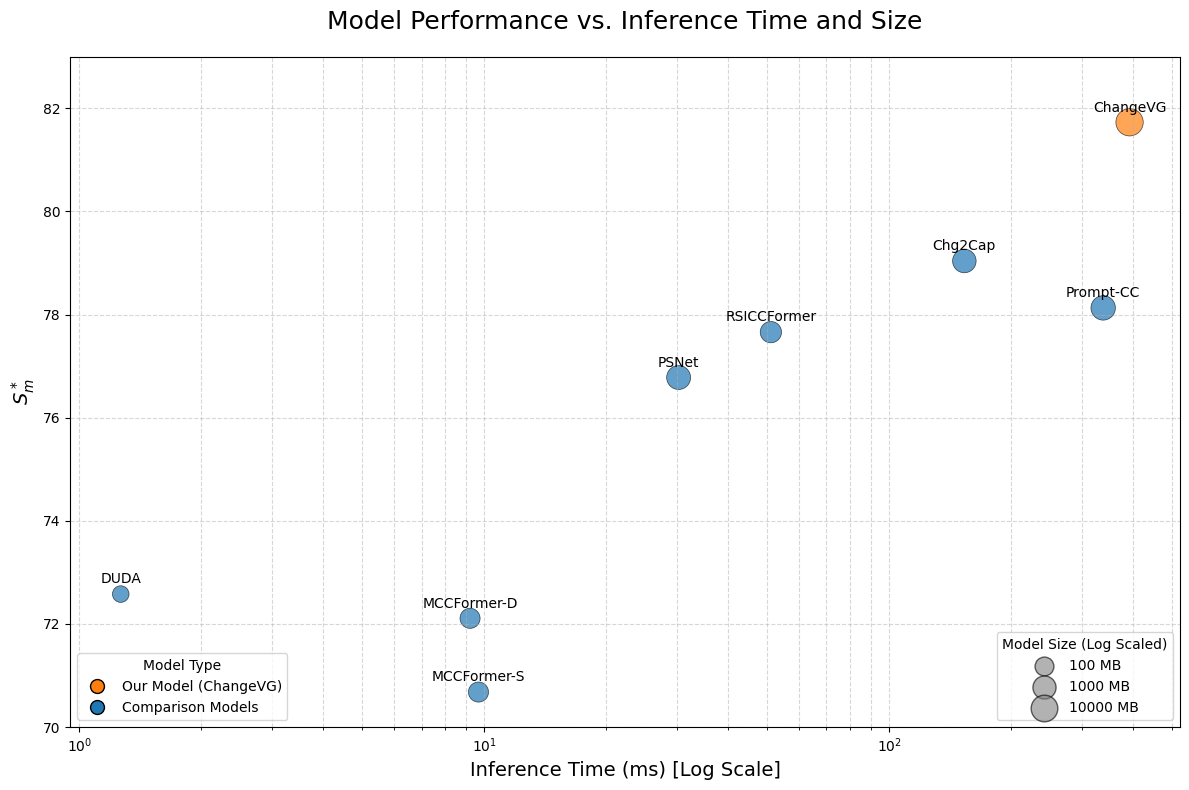}
    \caption{Comparison of model size, inference time, and model performance across different models. The x-axis represents inference time, the y-axis denotes model performance, and the circle size corresponds to model size.}
    \label{fig:size_time}
\end{figure}

We further investigate the trade-offs among model size, inference time, and model performance. As illustrated in Figure~\ref{fig:size_time}, this plot compares different models across three dimensions: the x-axis represents inference time on a logarithmic scale, the y-axis denotes model performance, and the size of the circles corresponds to model scale. Orange circles represent our model (ChangeVG), while blue circles denote comparative methods. As shown in Table~\ref{tab:size_time_table}, our approach outperforms other methods by up to 10\% and by at least 5\%, further underscoring its competitiveness in balancing these multidimensional trade-offs.

\begin{table*}[!t]
    \centering
    \caption{Comparison of model size, inference time, and model performance across different models.}
    \label{tab:size_time_table}
    \begin{tabular}{p{3.5cm} | P{2.2cm} P{2.2cm} P{2.2cm} P{2.2cm} }
        \toprule
         \textbf{Method} &  \textbf{Model Size(MB)} & \textbf{Inference time(ms)} & \textbf{$S_m^{*}$} & \textbf{Improvement to Ours}  \\
        \midrule
        DUDA ~\cite{DUDA}                        & 35.59 & 1.266 & 72.58 & -11.20\%  \\
        MCCFormer-S ~\cite{MCCFormer}            & 186.25 & 9.678 & 70.68 & -13.52\%  \\
        MCCFormer-D ~\cite{MCCFormer}            & 186.25 & 9.225 & 72.11 & -11.77\%  \\
        RSICCFormer \cite{RSICCFormer}           & 353.979 & 51.021 & 77.66 & -4.98\%  \\
        PSNet   \cite{PSNet}                     & 1666.85 & 30.198 & 76.78 & -6.01\%  \\
        Chg2Cap  \cite{Chg2Cap}                  & 1311.33 & 153.284 & 79.04 & -3.29\%  \\
        Prompt-CC  \cite{liu2023decoupling}      & 2315.34 & 337.496 & 78.13 & -4.40\%  \\
        \midrule
        ChangeVG (Ours)                          & 16285.25 & 392.121 & 81.73 & - \\
        \bottomrule
    \end{tabular}
\end{table*}

\subsubsection{Binary Change Classfication}
For this task, we compare our approach with several SOTA VLMs, such as GLM-4v~\cite{chatglm}, Qwen2.5-VL-7B , Qwen2.5-VL-72B~\cite{qwen2.5-vl}, GPT-4o~\cite{gpt-4o}, Gemini2.5-Flash~\cite{gemini} and Claude-Sonnet-4.0~\cite{claude}. All models are evaluated under the same setting: each receives a pair of bi-temporal images along with a binary instruction, and must classify whether a change is present.

\begin{table}[!t]
    \centering
    \caption{Performance comparisons with VLMs on the Binary Change Classification task.}
    \label{tab:change judge}
    \begin{tabular}{l | c c c c}
        \toprule
         \textbf{Method} &  \textbf{Accuracy} & \textbf{Precision} & \textbf{Recall} & \textbf{F1} \\
        \midrule
        GLM-4v \cite{chatglm}                   & 0.6580 & 0.6705 & 0.6679 & 0.6692 \\
        Qwen2.5-VL-7B  \cite{qwen2.5-vl}        & 0.5805 &  0.8888 & 0.2166  & 0.3483  \\
        Qwen2.5-VL-72B \cite{qwen2.5-vl}        & 0.7110 & \textbf{0.9711} & 0.4555 & 0.6202 \\
        GPT-4o    \cite{gpt-4o}                 & 0.7240  & 0.8558  & 0.5617 & 0.6783  \\
        Gemini2.5-Flash \cite{gemini}           & \underline{0.8036} & 0.7675 & \underline{0.8895}  &  \underline{0.8240} \\
        Claude-Sonnet-4.0  \cite{claude}        & 0.5170  & 0.8723 & 0.0791 & 0.1451 \\
        \midrule
        ChangeVG (Ours)                                & \textbf{0.9460} & \underline{0.9549} & \textbf{0.9401} & \textbf{0.9474} \\
        \bottomrule
    \end{tabular}
\end{table}

As shown in Table~\ref{tab:change judge}, our model can significantly outperform all the advanced baselines across four important evaluation metrics. In particular, our approach achieves an Accuracy of 94.6\% and an F1 score of 94.74\%, surpassing the best-performing baseline (Gemini2.5-Flash) by +14.26 and +12.34 points respectively. These results demonstrate the effectiveness of our visual-enhanced instruction tuning framework, which improves both precision and recall, and offers robust performance even in challenging change scenarios.

\begin{table}[!t]
    \centering
    \caption{Performance comparisons of MAE on Change Counting for Roads and Buildings across VLMs.}
    \label{tab:change count}
    \begin{tabular}{>{\raggedright\arraybackslash}p{3cm} | P{2cm} P{2cm}}
        \toprule
         \textbf{Method} &  \textbf{Road}~($\downarrow$) & \textbf{Build}~($\downarrow$)  \\
        \midrule
        GLM-4v \cite{chatglm}                 & 0.486 & 4.161 \\
        Qwen2.5-VL-7B \cite{qwen2.5-vl}         & 0.943 & 3.875 \\
        Qwen2.5-VL-72B \cite{qwen2.5-vl}         & 0.408 & 4.181 \\
        GPT-4o \cite{gpt-4o}                   & \underline{0.348} & \underline{3.663} \\
        Gemini2.5-Flash  \cite{gemini}      & 1.295 & 3.909 \\
        Claude-Sonnet-4.0 \cite{claude}     & 0.711 & 4.308 \\
        \midrule
        ChangeVG (Ours)               & \textbf{0.156} & \textbf{0.802}\\
        \bottomrule
    \end{tabular}
\end{table}

\begin{table*}[!t]
    \centering
    \caption{Performance comparisons with VLM on the Road Change Localization.}
    \label{tab:change-location-1}
    \resizebox{\textwidth}{!}{
    \begin{tabular}{l | l | c c c c c}
        \toprule
         \multicolumn{2}{c}{\textbf{Method}} & \textbf{Example-based Accuracy} &  \textbf{Micro Precision} & \textbf{Micro Recall} & \textbf{Micro F1} & \textbf{Subset Accuracy}  \\
        \midrule
         \multirow{8}{*}{Road}
        & GLM-4v \cite{chatglm}                 & 0.5451 & 0.3659 & 0.5881 & 0.4511 & 0.5030 \\
        & Qwen2.5-VL-7B \cite{qwen2.5-vl}         & 0.0088 & 0.0219 & 0.0108 & 0.0145 & 0.0060 \\
        & Qwen2.5-VL-72B \cite{qwen2.5-vl}        & 0.4746 & 0.4771 & 0.3993 & 0.4347 & 0.4720 \\
        & GPT-4o   \cite{gpt-4o}               & 0.6010 & 0.4724 & 0.6076 & 0.5315 & 0.5608 \\
        & Gemini2.5-Flash \cite{gemini}       & 0.2712 & 0.1579 & \underline{0.6667} & 0.2554 & 0.1800 \\
        & Claude-Sonnet-4.0 \cite{claude}     & \underline{0.6475} & \underline{0.4789} & 0.6578 & \underline{0.5543} & \underline{0.6338} \\
        \midrule
        & ChangeVG (Ours)              & \textbf{0.7865} & \textbf{0.7850} & \textbf{0.7595} & \textbf{0.7720} & \textbf{0.7330} \\
        \bottomrule
    \end{tabular}
    }
\end{table*}

\begin{table*}[!t]
    \centering
    \caption{Performance comparisons with VLMs on the Building Change Localization.}
    \label{tab:change-location-2}
    \resizebox{\textwidth}{!}{
    \begin{tabular}{l | l | c c c c c}
        \toprule
         \multicolumn{2}{c}{\textbf{Method}} & \textbf{Example-based Accuracy} &  \textbf{Micro Precision} & \textbf{Micro Recall} & \textbf{Micro F1} & \textbf{Subset Accuracy}  \\
        \midrule
        \multirow{8}{*}{Building} & GLM-4v \cite{chatglm}            & 0.6093 & 0.6295 & 0.4619 & 0.5328 & 0.5310 \\
        & Qwen2.5-VL-7B  \cite{qwen2.5-vl}        & 0.0460 & 0.2012 & 0.0483 & 0.0770 & 0.0140  \\
        & Qwen2.5-VL-72B  \cite{qwen2.5-vl}       & 0.4974 & 0.5355 & 0.2047 & 0.2962 & 0.4800  \\
        & GPT-4o    \cite{gpt-4o}              & \underline{0.6652} & \underline{0.7766} & 0.4293 & 0.5530 & \underline{0.5872}  \\
        & Gemini2.5-Flash  \cite{gemini}      & 0.6012 & 0.6353 & \underline{0.8285} & \underline{0.7192} & 0.3950  \\
        & Claude-Sonnet-4.0 \cite{claude}     & 0.5318 & 0.3828 & 0.5134 & 0.4386 & 0.4889  \\ 
        \midrule
        & ChangeVG (Ours)              & \textbf{0.8813} & \textbf{0.9068} & \textbf{0.9208} & \textbf{0.9137} &  \textbf{0.7220}\\
        \bottomrule
    \end{tabular}
    }
\end{table*}

\subsubsection{Change Counting}
As shown in Table~\ref{tab:change count}, our model significantly outperforms all baseline VLMs. To be specific, our method achieves an MAE of 0.156 for roads and 0.802 for buildings, outperforming the best-performing baseline (GPT-4o for roads with 0.348 and GPT-4o for buildings with 3.663) by a large margin.

These improvements demonstrate the effectiveness of our design, where the fine-grained recognition branch explicitly segments and localizes change regions, and the visual auxiliary module enriches the model’s understanding of spatial details. This fine-grained guidance allows the VLMs to make more accurate numerical predictions during instruction-following, particularly in dense or small-scale change scenarios.

\subsubsection{Change Localization}
As displayed in Tables~\ref{tab:change-location-1} \& \ref{tab:change-location-2}, our method significantly outperforms existing VLMs across all the evaluation metrics. These metrics can jointly reflect both the overall correctness of model predictions (Precision and Recall) and the capability to fully capture multi-label outputs (\textit{i.e.}, subset accuracy).

For the road category, our model achieves a Micro F1 of 0.7720, surpassing the best baseline (Claude-Sonnet-4.0, 0.5543) by 21.77 percentage points. Similarly, for the building, our approach achieves 0.9137, improving upon the second-best result (Gemini2.5-Flash, 0.7192) by a large margin.

The superior performance can be attributed to the fine-grained recognition branch, which explicitly extracts and segments multi-scale spatial features, and the visual auxiliary encoder, which enhances semantic grounding in the image domain. Together, these components provide strong visual priors to the language model, enabling it to localize change regions with high accuracy.

\begin{figure*}[!t]
    \centering
    \includegraphics[width=\linewidth]{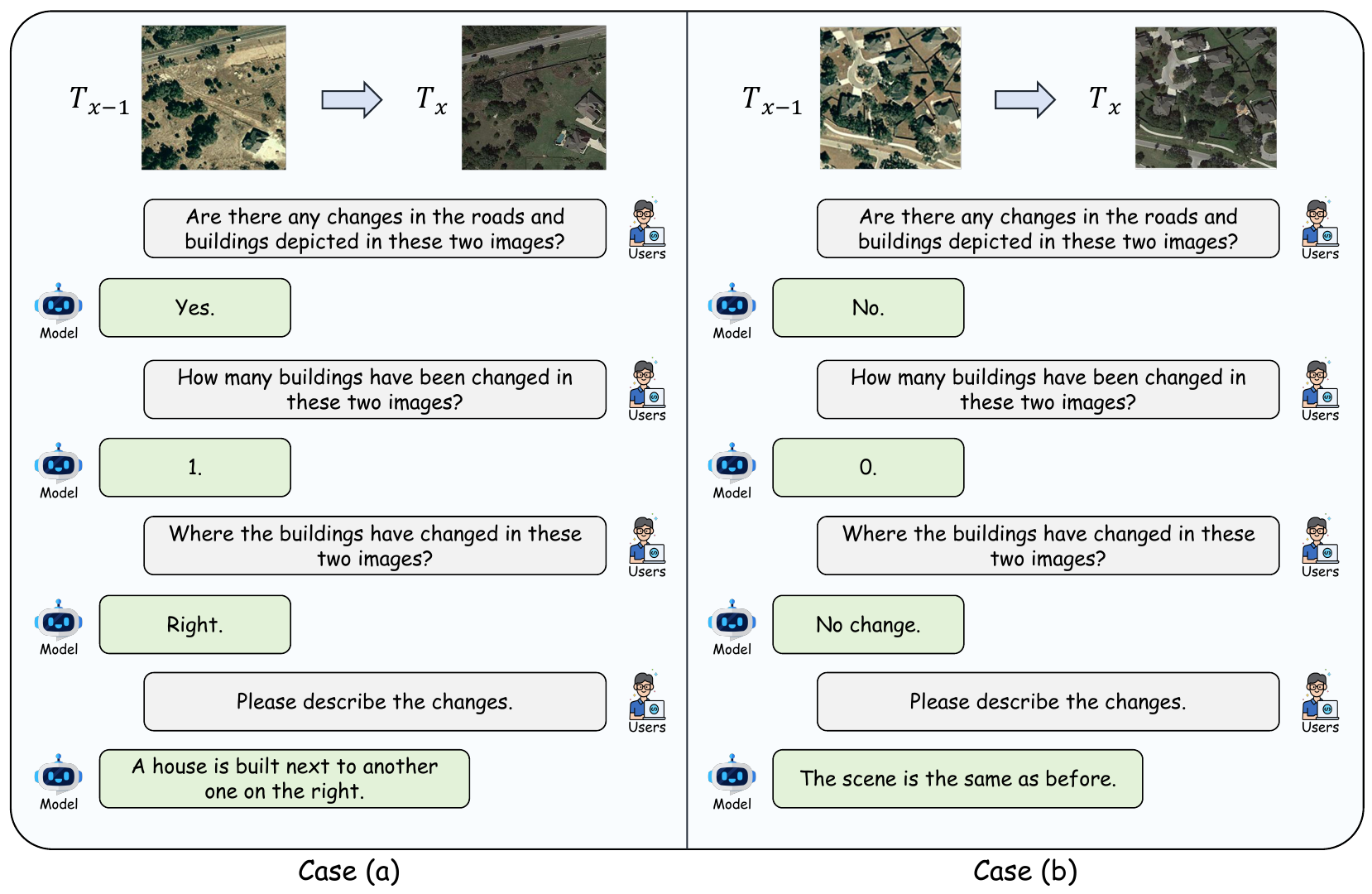}
    \vspace{-1.5em}
    \caption{The conversations example of multi-turn with different user queries. The $T_{x-1}$ and $T_x$ represent images from two different moments in time, with $T_{x-1}$ being the image from the earlier time and $T_x$ being the one from the later time. Case (a) illustrates a scenario with changes, while Case (b) illustrates a scenario without changes.}
    \vspace{-1.3em}
    \label{fig:mul}
\end{figure*}

In addition to single-turn evaluations, we design task-oriented multi-turn dialogue settings to assess the model’s capability to engage in coherent, step-by-step reasoning across temporally aligned image pairs, while also demonstrating its ability to interact with users. As shown in Fig.~\ref{fig:mul}.

\subsection{Ablation Studies}

\begin{table*}[!t]
    \centering
    \renewcommand{\arraystretch}{0.90}
    \caption{The effects of auxiliary tasks and Vision-guided on change captioning performance.}
    \label{tab:Ablation_Studies_1}
    \resizebox{\textwidth}{!}{
    \begin{tabular}{c c c c | c c c c | c}
        \toprule
         \textbf{Caption} & \textbf{Count + Location} & \textbf{Binary} & \textbf{Vision-guided} & \textbf{BLEU-4} & \textbf{METEOR} & \textbf{ROUGE$_L$} & \textbf{CIDEr-D} & \textbf{$S_m^{*}$} \\
        \midrule
        $\checkmark$ & $\times$ & $\times$ & $\times$           & 61.30 & 40.69 & 74.61 & 138.08 & 78.67 \\
        $\checkmark$ & $\times$ & $\checkmark$ & $\times$       & 61.00 & 40.15 & 74.46 & 137.52 & 78.28 \\
        $\checkmark$ & $\checkmark$ & $\times$ & $\times$       & 61.79 & 40.69 & 74.93 & 139.23 & 79.16 \\
        $\checkmark$ & $\checkmark$ & $\checkmark$ & $\times$   & \underline{62.98} & \underline{41.11} & \underline{75.79} & \underline{140.48} & \underline{80.09} \\
        \midrule
        $\checkmark$ & $\checkmark$ & $\checkmark$ & $\checkmark$              & \textbf{65.08} & \textbf{42.06} & \textbf{76.95} & \textbf{142.83} & \textbf{81.73} \\
        \bottomrule
    \end{tabular}
    }
    \vspace{0.3em}
\end{table*}

\subsubsection{Diffence task and vision-guided}
To further investigate the contributions of individual auxiliary tasks and the vision-guided module to the overall performance of our model on the change captioning task, we conduct detailed ablation studies. As shown in Table~\ref{tab:Ablation_Studies_1}, we progressively introduce binary change classification, change counting and change localization tasks, as well as the vision-guided  module into the instruction tuning process, and evaluate their respective impacts.

\begin{figure*}[!t]
    \centering
    \includegraphics[width=\linewidth]{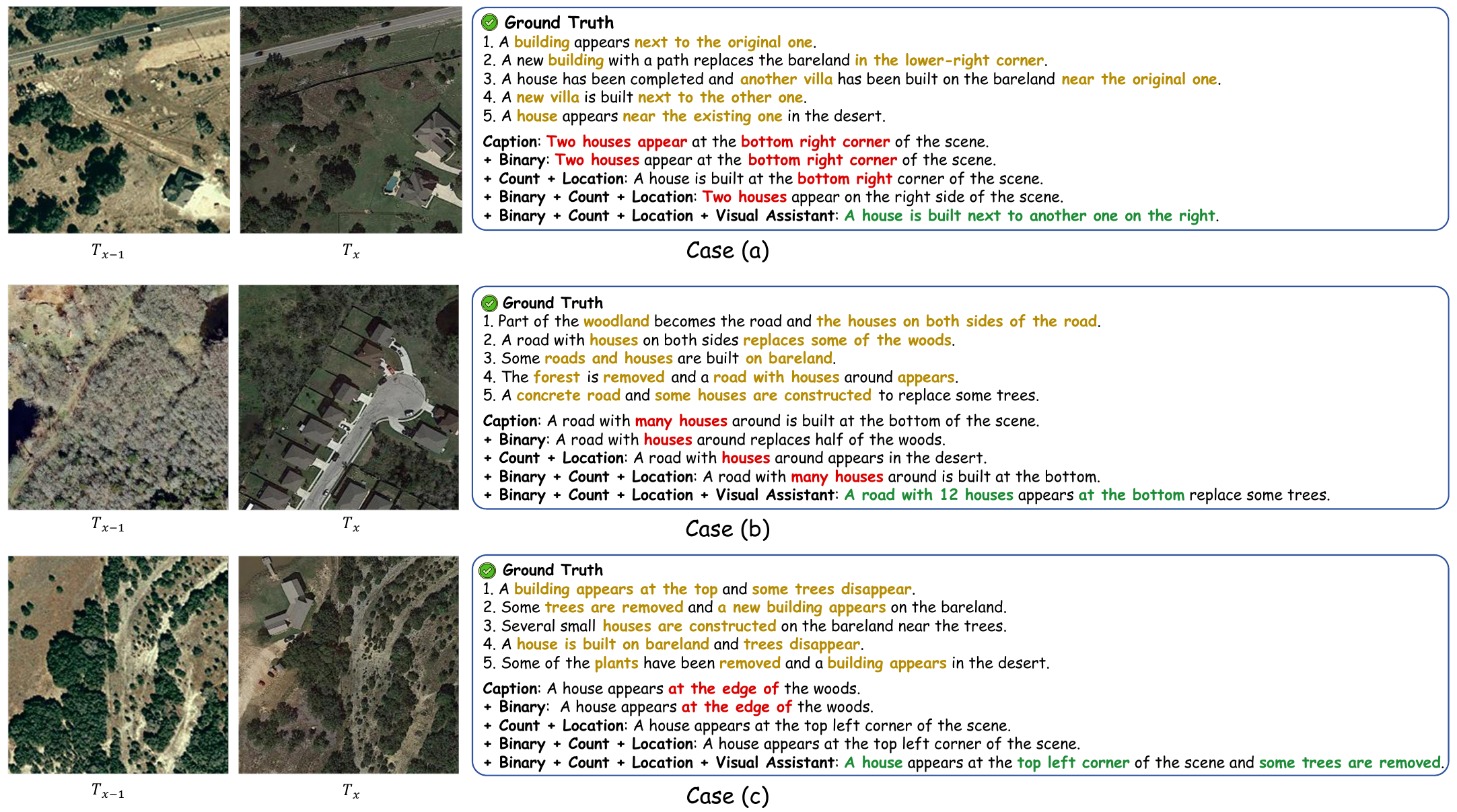}
    \caption{Example from the change captioning task used for the ablation study. Texts in green indicate correct descriptions, yellow highlights represent partially accurate or ambiguous expressions, and red indicates incorrect outputs. }
    \label{fig:xiaorong}
\end{figure*}

\begin{table*}[!t]
    \centering
    \caption{The practicality of the Vision-guided  module on other VLMs.}
    \label{tab:Ablation_Studies_2}
    \resizebox{\textwidth}{!}{
    \begin{tabular}{p{3.5cm} | P{2.2cm} P{2.2cm} P{2.2cm} P{2.2cm} | P{2.2cm}}
        \toprule
         \textbf{Method} & \textbf{BLEU-4} & \textbf{METEOR} & \textbf{ROUGE$_L$} & \textbf{CIDEr-D} & \textbf{$S_m^{*}$} \\
        \midrule
        InternVL3-8B ~\cite{internvl3}              & 65.89 & 42.09 & 76.49 & 142.50 & 81.74 \\
        InternVL3-8B + visual       & 63.17 & 42.69 & 76.95 & 145.49 & 82.08 \\
        \midrule
        GLM-4.1v-9B ~\cite{glm4.1}             & 61.77 & 41.43 & 75.45 & 140.01 & 79.67 \\
        GLM-4.1v-9B + visual        & 64.33 & 41.81 & 76.27 & 141.04 & 80.86 \\
        \bottomrule
    \end{tabular}
    }
    \vspace{0.3em}
\end{table*}

We first consider the single-task baseline that trains only on the change captioning objective. This configuration achieves a CIDEr-D of 138.08 and an $S_m^*$ score of 78.67. Interestingly, when introducing the binary change classification task alone, the model’s performance slightly decreases across multiple metrics (\textit{e.g.}, CIDEr-D drops to 137.52, METEOR drops to 40.15). This suggests that binary classification, being a coarse-grained task, may not provide sufficiently rich semantic signals to support fine-grained language generation and might even introduce conflicting gradients during optimization.

In contrast, integrating the change counting and change localization tasks alone yields clear performance improvements (CIDEr-D rises to 139.23, $S_m^*$ to 79.16), demonstrating that spatially and numerically grounded auxiliary tasks offer more aligned and informative supervision for descriptive caption generation. When all three auxiliary tasks are jointly trained, the model benefits further, achieving a CIDEr-D of 140.48 and an $S_m^*$ score of 80.09, indicating that the combination of coarse and fine-grained tasks enhances the model’s overall semantic understanding.

In the end, the incorporation of the proposed vision-guided module brings the most significant performance gains. Equipped with this module, the model attains the best results across all metrics (\textit{e.g.}, CIDEr-D of 142.83 and $S_m^*$ of 81.73), confirming that integrating structured visual guidance can effectively enhance the model’s ability to perceive and describe nuanced bi-temporal changes. These results collectively validate the effectiveness of our multi-task and vision-augmented training strategy in producing accurate and context-aware change captions. As shown in Fig.~\ref{fig:xiaorong}, we present the visualization results of the captioning task under different ablation settings. It can be clearly observed that without the vision-guided module, the generated captions exhibit noticeable inaccuracies in both quantity estimation and spatial localization of the changes.

\subsubsection{fine-grained recognition branch and global summary branch}
\begin{table*}[!t]
    \centering
    \caption{The effects of fine-grained recognition branch and global summary branch on the performance of change captioning. w/o refers to without.}
    \label{tab:Ablation_Studies_3}
    \begin{tabular}{p{3.5cm} | P{2.2cm} P{2.2cm} P{2.2cm} P{2.2cm} | P{2.2cm}}
        \toprule
         \textbf{Method} &  \textbf{BLEU-4} & \textbf{METEOR} & \textbf{ROUGE$_L$} & \textbf{CIDEr-D} & \textbf{$S_m^{*}$} \\
        \midrule
        w/o vision-guided                              & 62.98 & 41.11 & 75.79 & 140.48 & 80.09 \\
        w/o fine-grained recognition                   & 63.92 & 42.64 & 76.86 & 142.02 & 81.36 \\
        w/o global summary                             & 62.12 & 42.11 & 76.37 & 142.41 & 80.75 \\
        \midrule
        ChangeVG(all)                                  & \textbf{65.08} & \textbf{42.06} & \textbf{76.95} & \textbf{142.83} & \textbf{81.73} \\
        \bottomrule
    \end{tabular}
\end{table*}

\begin{table}[!t]
    \centering
    \caption{The effects of fine-grained recognition branch and global summary branch on binary change classification performance.}
    \label{tab:Ablation_Studies_class}
    \begin{tabular}{l | c c c c}
        \toprule
         \textbf{Method} &  \textbf{Accuracy} & \textbf{Precision} & \textbf{Recall} & \textbf{F1} \\
        \midrule
        w/o vision-guided                  & 0.9220  & 0.9246  & 0.9209  & 0.9228  \\
        w/o Fine-Grained                  & 0.9420  & 0.9618  & 0.9190  & 0.9399 \\
        w/o Global                         & 0.9440  & 0.9620  & 0.9231  & 0.9421 \\
        \midrule
        ChangeVG (Ours)                                & \textbf{0.9460} & \underline{0.9549} & \textbf{0.9401} & \textbf{0.9474} \\
        \bottomrule
    \end{tabular}
\end{table}

\begin{table}[!t]
    \centering
    \caption{The effects of fine-grained recognition branch and global summary branch on change counting performance.}
    \label{tab:Ablation_Studies_count}
    \begin{tabular}{>{\raggedright\arraybackslash}p{3cm} | P{2cm} P{2cm}}
        \toprule
         \textbf{Method} &  \textbf{Road}~($\downarrow$) & \textbf{Build}~($\downarrow$)  \\
        \midrule
        w/o vision-guided                & 0.164 & 0.860 \\
        w/o Fine-Grained                 & 0.161 & 0.855 \\
        w/o Global                       & 0.158 & 0.811 \\
        \midrule
        ChangeVG (Ours)               & \textbf{0.156} & \textbf{0.802}\\
        \bottomrule
    \end{tabular}
\end{table}

\begin{table*}[!t]
    \centering
    \caption{The effects of fine-grained recognition branch and global summary branch on road change localization.}
    \label{tab:Ablation_Studies_location_road}
    \resizebox{\textwidth}{!}{
    \begin{tabular}{l | l | c c c c c}
        \toprule
         \multicolumn{2}{c}{\textbf{Method}} & \textbf{Example-based Accuracy} &  \textbf{Micro Precision} & \textbf{Micro Recall} & \textbf{Micro F1} & \textbf{Subset Accuracy}  \\
        \midrule
         \multirow{4}{*}{Road}
        & w/o vision-guided                 & 0.7697 & 0.7688 & 0.7513 & 0.7600 & 0.7056 \\
        & w/o Fine-Grained                  & 0.7719 & 0.7711 & 0.7491 & 0.7599 & 0.7076 \\
        & w/o Global                        & 0.7843 & 0.7813 & 0.7618 & 0.7714 & 0.7260 \\
        \midrule
        & ChangeVG (Ours)              & \textbf{0.7865} & \textbf{0.7850} & \textbf{0.7595} & \textbf{0.7720} & \textbf{0.7330} \\
        \bottomrule
    \end{tabular}
    }
\end{table*}

\begin{table*}[!t]
    \centering
    \caption{The effects of fine-grained recognition branch and global summary branch on building change localization.}
    \label{tab:Ablation_Studies_location_building}
    \resizebox{\textwidth}{!}{
    \begin{tabular}{l | l | c c c c c}
        \toprule
         \multicolumn{2}{c}{\textbf{Method}} & \textbf{Example-based Accuracy} &  \textbf{Micro Precision} & \textbf{Micro Recall} & \textbf{Micro F1} & \textbf{Subset Accuracy}  \\
        \midrule
        \multirow{4}{*}{Building} 
        & w/o vision-guided          & 0.8547 & 0.9105 & 0.9038 & 0.9071 & 0.6580  \\
        & w/o Fine-Grained           & 0.8649 & 0.9132 & 0.9158 & 0.9145 & 0.6817  \\
        & w/o Global                 & 0.8784 & 0.9181 & 0.9105 & 0.9142 & 0.7119  \\ 
        \midrule
        & ChangeVG (Ours)            & \textbf{0.8813} & \textbf{0.9068} & \textbf{0.9208} & \textbf{0.9137} &  \textbf{0.7220}\\
        \bottomrule
    \end{tabular}
    }
\end{table*}

To evaluate the effectiveness of the fine-grained recognition branch and the global summary branch in our Vision-guided Module, we conducted ablation studies on each branch individually. Specifically, we performed  experiments by masking the information from the respective branch within the vision-guided module. As shown in table~\ref{tab:Ablation_Studies_3}, both branches contribute to improved performance on the captioning task. Notably, since captioning is inherently a global-level task, the model benefits more from the global summary branch alone than from the fine-grained recognition branch alone. We also evaluated the performance of the two branches on these three additional tasks.

As shown in Table~\ref{tab:Ablation_Studies_class}, both branches improve the model’s performance on binary change classification. However, since this task is relatively simple, the contributions of the two branches are comparable. As shown in Table~\ref{tab:Ablation_Studies_count}, the two branches exhibit distinct contributions on the change counting task. Since this task is fine-grained in nature, the fine-grained recognition branch provides a greater performance gain compared to the global summary branch. Moreover, combining both branches, which integrate global and local information, leads to further performance improvement. 
Table~\ref{tab:Ablation_Studies_location_road} and table~\ref{tab:Ablation_Studies_location_building} present the performance of the two branches on the change localization tasks for roads and buildings, respectively. Since these tasks are also fine-grained in nature, the fine-grained recognition branch contributes more significantly to performance improvement than the global summary branch.

\subsubsection{Vision-guided Module in diffent VLMs}
To further verify the applicability of our proposed vision-guided module, we conducted additional comparative experiments on two open-source VLMs, InternVL3-8B~\cite{internvl3} and GLM-4.1v-9B~\cite{glm4.1}. Specifically, we first fine-tuned the baseline models on the ChangeIMTI dataset, and then further fine-tuned them by incorporating the vision-guided module to assess the improvements brought by this module. As shown by table ~\ref{tab:Ablation_Studies_2}, after incorporating the vision-guided module, all metrics on the captioning task improved, demonstrating the module’s ability to enhance natural language understanding. For instance, the $S^*_m$ for InternVL3-8B increased from 81.74 to 82.08, and for GLM-4.1v-9B, it increased from 79.67 to 80.86.

\subsection{Zero-shot Evaluation}

\begin{table}[!t]
    \centering
    \caption{Performance Comparisons with VLMs on Binary Change Classification in QAG-360k}
    \label{tab:zero_shot_qag}
    \begin{tabular}{l | c c c c}
        \toprule
         \textbf{Method} &  \textbf{Accuracy} & \textbf{Precision} & \textbf{Recall} & \textbf{F1} \\
        \midrule
        GLM-4v \cite{chatglm}                   &  \underline{0.7677} & 0.6667 & 0.7894 & \underline{0.7229} \\
        Qwen2.5-VL-7B  \cite{qwen2.5-vl}        &  0.6970 &  0.5714 & 0.6667  & 0.6153  \\
        Qwen2.5-VL-72B \cite{qwen2.5-vl}        & 0.7349 & \textbf{0.8846} & 0.4339 & 0.5822 \\
        GPT-4o    \cite{gpt-4o}                 & 0.6566  & 0.5322  & 0.8684 & 0.6600  \\
        Gemini2.5-Flash \cite{gemini}           & \underline{0.7677} & 0.6829 & 0.7368  &  0.7089 \\
        Claude-Sonnet-4.0  \cite{claude}        & 0.5556  & 0.4595 &  \textbf{0.8947} & 0.6071 \\
        \midrule
        ChangeVG (Ours)                                & \textbf{0.8081} & \underline{0.7111} & \underline{ 0.8421} & \textbf{0.7711} \\
        \bottomrule
    \end{tabular}
\end{table}

\begin{table}[!t]
    \centering
    \caption{Performance Comparisons with VLMs on Binary Change Classification in CDVQA}
    \label{tab:zero_shot_cd}
    \begin{tabular}{l | c c c c}
        \toprule
         \textbf{Method} &  \textbf{Accuracy} & \textbf{Precision} & \textbf{Recall} & \textbf{F1} \\
        \midrule
        GLM-4v \cite{chatglm}                   &   0.7475 & 0.8113 & 0.7414 & 0.7748 \\
        Qwen2.5-VL-7B  \cite{qwen2.5-vl}        &  0.5859 &  0.8400 & 0.3621  & 0.5060  \\
        Qwen2.5-VL-72B \cite{qwen2.5-vl}        &  0.7727 & \textbf{0.8889} & 0.6667 & 0.7619 \\
        GPT-4o    \cite{gpt-4o}                 & 0.7273  & 0.7067  & \textbf{0.9138} & \underline{0.7970}  \\
        Gemini2.5-Flash \cite{gemini}           & 0.6465 & 0.7674 & 0.5690  &   0.6535 \\
        Claude-Sonnet-4.0  \cite{claude}        & 0.5960  & 0.6286 &  0.7586 & 0.6875 \\
        \midrule
        ChangeVG (Ours)                                & \textbf{0.8182} & \underline{0.8704} & \underline{0.8103} & \textbf{0.8393} \\
        \bottomrule
    \end{tabular}
\end{table}

To further validate the generalization capability of our introduced ChangeVG, we conduct zero-shot experiments on two publicly available remote sensing change understanding datasets that demonstrate substantial overlap with ChangeIMTI: QAG-360k~\cite{QAG} and CDVQA~\cite{CDVQA}. 

QAG-360 contains 10 classes, 6,810 image pairs, and 8 tasks, including the "change or not" task. This aligns directly with our binary change classification task, as both aim to determine whether a change occurs between image pairs.
CDVQA comprises 30 change categories, 4,662 image pairs, and 8 tasks, including a "change or not" task equivalent to our binary change classification task.
To maintain consistency with our dataset's task definition, we evaluate the performance of binary change classification on these two datasets and compare it with several SOTA VLMs. 

Table~\ref{tab:zero_shot_qag} and Table~\ref{tab:zero_shot_cd} present the evaluation results of our model alongside other SOTA VLMs on the QAG-360k and CDVQA datasets. As shown, our model consistently outperforms competing approaches in both accuracy and F1 score, demonstrating its superior generalization capability in the domain of remote sensing change understanding.

\section{Conclusion and Future Works}
In this paper, we present a unified framework ChangeVG for RSCU built upon VLMs. To facilitate instruction tuning across diverse tasks, we construct ChangeIMTI, a large-scale interactive multi-task dataset that includes change captioning, binary change classification, object counting, and change localization, to evaluate VLMs in remote sensing.
Our framework incorporates a vision-guided module, which enhances the model’s ability to capture both fine-grained and global changes across bi-temporal images. Through this integration, the model supports both caption generation and VQA tasks, enabling semantically interpretable and task-aware reasoning. At the same time, our model is capable of interactive question answering with users, providing precise answers about RSCU based on specific questions posed by the user. 
Extensive experiments conducted on multiple RSCU tasks demonstrate that our approach achieves SOTA performance, validating both the effectiveness of our unified design and its generalizability across tasks. The work highlights the potential of VLMs in remote sensing applications and provides a solid foundation for future research in vision-language geospatial understanding.

Despite promising results, the current ChangeIMTI contains a limited number of object categories, which may constrain the model's ability to generalize to more diverse scenes. In the future work, we plan to expand the dataset to include a broader range of object types, thereby enhancing the model's generalization capability. In addition, incorporating Chain-of-Thought (CoT) pattern into the training samples can be explored to further improve the explainability of model predictions.

\bibliographystyle{IEEEtran}
\bibliography{main} 

\end{document}